\pdfoutput=1
\documentclass[runningheads]{llncs}
\usepackage{graphicx}
\usepackage{comment}
\usepackage{hyperref}
\usepackage{url}
\usepackage{epsfig}
\usepackage{mathtools,nccmath}
\usepackage{physics}
\usepackage{multirow}
\usepackage{subfig}
\usepackage{amsmath,amssymb} 
\usepackage{color}
\usepackage{layouts}

\begin{document}
\pagestyle{headings}
\mainmatter
\def\ECCVSubNumber{4098}  

\title{Why Are Deep Representations Good Perceptual Quality Features?} 

\titlerunning{Why Are Deep Representations Good Perceptual Quality Features?}
%
\author{Taimoor Tariq\inst{1} \and
Okan Tarhan Tursun\inst{1} \and
Munchurl Kim\inst{2} \and
Piotr Didyk\inst{1}}
\authorrunning{T. Tariq et al.}
%
\institute{Università della Svizzera italiana, Switzerland. \and
KAIST, South Korea.
}
\maketitle

\begin{abstract}
Recently, intermediate feature maps of pre-trained convolutional neural networks have shown significant perceptual quality improvements, when they are used in the loss function for training new networks. It is believed that these features are better at encoding the perceptual quality and provide more efficient representations of input images compared to other perceptual metrics such as SSIM and PSNR. However, there have been no systematic studies to determine the underlying reason. Due to the lack of such an analysis, it is not possible to evaluate the performance of a particular set of features or to improve the perceptual quality even more by carefully selecting a subset of features from a pre-trained CNN. This work shows that the capabilities of pre-trained deep CNN features in optimizing the perceptual quality are correlated with their success in capturing basic human visual perception characteristics. In particular, we focus our analysis on fundamental aspects of human perception, such as the contrast sensitivity and orientation selectivity. We introduce two new formulations to measure the frequency and orientation selectivity of the features learned by convolutional layers for evaluating deep features learned by widely-used deep CNNs such as VGG-16. We demonstrate that the pre-trained CNN features which receive higher scores are better at predicting human quality judgment. Furthermore, we show the possibility of using our method to select deep features to form a new loss function, which improves the image reconstruction quality for the well-known single-image super-resolution problem.                                      


\end{abstract}

\section{Introduction}
\label{intro}
The loss functions based on features from deep convolutional neural networks (CNNs) pre-trained for image classification have been shown to correlate well with human quality perception and have successful applications in image processing problems \cite{12}. The perceptual loss proposed by Johnson~et~al.~\cite{7} was one of the first studies which showed how effective the distance between feature representations of pre-trained CNNs could be for improving perceptual quality, especially when they are used in loss functions for training other networks. The effectiveness of the loss functions based on deep CNN representations has been further demonstrated in more recent works \cite{12,5,32}. Consequently, the perceptual loss is now widely used in many common image enhancement and reconstruction tasks such as super-resolution, style transfer, denoising, etc. \cite{9,10,11,TT2019}. Unfortunately, the analysis of underlying characteristics of the deep features as well as the explanation of their superior performance in quantifying visual distortions are still incomplete. 

Classical models that predict the magnitude of the perceived difference between images rely on models of the human visual system (HVS). They are usually easy to interpret, evaluate, and fine-tune as necessary. On the other hand, neural networks are mostly used as non-linear black-boxes with little intuition on the process leading to their output, which makes a tractable analysis nontrivial. Most of our understanding of visual quality perception is obtained from psychophysical experiments that investigate basic phenomena such as the effect of spatial frequency and orientation of stimuli on perception. Those studies resulted in well-known HVS models, such as the contrast sensitivity function (CSF), which allow us to perceptually quantify visual stimuli and visibility of differences \cite{33}. In this work, we investigate how visual information is encoded by pre-trained CNN features and take a step towards explaining the remarkable success of those features in improving the perceptual quality when they are used in the loss function for training new CNNs. We focus our analysis on two fundamental properties of the HVS, namely, the contrast sensitivity and orientation selectivity. In order to quantify the frequency and orientation selectivity of different channels in pre-trained CNNs, we compute the hidden intermediate network features for input image patches of synthesized sinusoidal gratings with a wide range of spatial frequencies and orientations. We then formulate two measures of the spatial frequency and orientation selectivity of feature channels based on mean channel activations. This allows us to analyze the role of those two perceptual attributes and how they are encoded in the network. Although HVS quality perception is a complex process and it is not limited to spatial contrast frequency and orientation perception, these two attributes play an important role in driving quality perception by determining visibility based on frequency characteristics of image distortions and structural differences. Consequently, these two attributes are the foundation of many classical models \cite{34,35,51,52,53}. The main hypothesis of this work is that in a pre-trained convolutional layer, the channels that share more similarities with the human CSF and those that offer better orientation selectivity are more useful for optimizing perceptual quality compared to the other channels in the layer. 

We verify our hypothesis using standard subjective tests such as quality assessment (QA test) \cite{24}, two-alternative forced choice (2AFC) experiments, and just-noticeable difference (JND) tests, which are psychophysical experiment protocols designed for measuring the correlation of visual quality predictions with human assessment. We rank and group the channels in different CNN layers into subsets according to our frequency and orientation selectivity metric scores and demonstrate that the groups of features with higher metric scores provide better perceptual quality. We repeat our experiment across multiple layers of different pre-trained image classification networks such as the VGG-16 \cite{30}, AlexNet \cite{1}, ShuffleNet \cite{41} and SqueezeNet \cite{36}. We demonstrate that using very large feature sets motivated by the goal of having a comprehensive representation of data is not necessarily a good practice. Instead, it may lead to quality degradation due to the inherent redundancy of features. We also analyze the effect of feature selection on the performance of calibrated and commonly used LPIPS metric in JND, 2AFC, and mean-opinion-scores MOS correlation tests \cite{12}.

\section{Deep CNN Representations as Perceptual Quality Features}
\label{def}

The visual quality obtained using deep learning solutions for image processing tasks, such as super-resolution or style transfer, is primarily driven by the particular loss function used during training. The performance of those solutions is mostly defined by the visual quality as perceived by a human observer. Simple per-pixel loss functions do not optimize the perceived quality because they do not resemble complex neurological and cognitive processes of the HVS. As a result, a direct comparison of pixel values in the loss function yields sub-optimal results. A better solution is to use feature maps from deep CNNs that are pre-trained for image classification. These maps represent images transformed in a higher-dimensional space that is more closely related to the processing performed by the HVS. The resulting distance measure, so-called \emph{perceptual loss} $\mathcal{L}_\text{p}$, between two images, $ I_1 $ and $ I_2 $, is usually defined as:
%
%
\begin{equation}
\label{eq:1}
\mathcal{L}_\text{p}^k \left( I_1, I_2 \right) = \frac{1}{M \cdot H \cdot W}\sum\limits_{m=1}^M\Vert\Phi^k_m(I_1)-\Phi^k_m(I_2)\Vert_2^2,
\end{equation}
where $ \Phi^k_m ( \cdot ) $ is the feature map from $m^{th}$ channel in $k^{th}$ convolutional layer of the particular deep CNN used. $M$ is the total number of channels where the output of each channel is an $H \times W$ feature map. In practice, training a network by minimizing the loss $ \mathcal{L}_\text{p}^k \left( I_{\text{out}}, I_{\text{GT}} \right) $ between the output, $ I_{\text{out}} $, and the ground truth, $ I_{\text{GT}} $, improves the correlation with human quality judgments.

\section{Problem Formulation}
\label{prob_def}

The use of loss functions based on deep CNN representations, as defined in Eq.~\ref{eq:1}, has been remarkably successful for optimizing perceived quality in the past. However, the source of this success is not analyzed from the visual perception perspective. Also, some feature channels perform better than others. These observations bring two important questions. First, why are some channels within a layer perform better than the others, and can we establish a connection between visual mechanisms involved in human perception and those features? Second, can we use the insight gained from establishing such a connection to rank the feature channels and carefully select a better subset by eliminating the redundant feature channels that correlate poorly with human perception? 


Establishing a connection between human visual perception and CNN representations is difficult because of the `black-box' nature of neural networks. In Sec.~\ref{sec:method}, we introduce a methodology to quantify the spatial frequency and orientation tuning of channels in pre-trained CNNs. Using this formulation, we interpret and explain deep CNN features as perceptual quality features by using basic human visual perception models, which rely on spatial frequency and orientation characteristics of input stimuli. In essence, the formulation in Sec.~\ref{sec:method} acts as a bridge between attributes of deep representations and fundamental visual perception properties. 

\begin{figure*} [t!]
\centering
\includegraphics[]{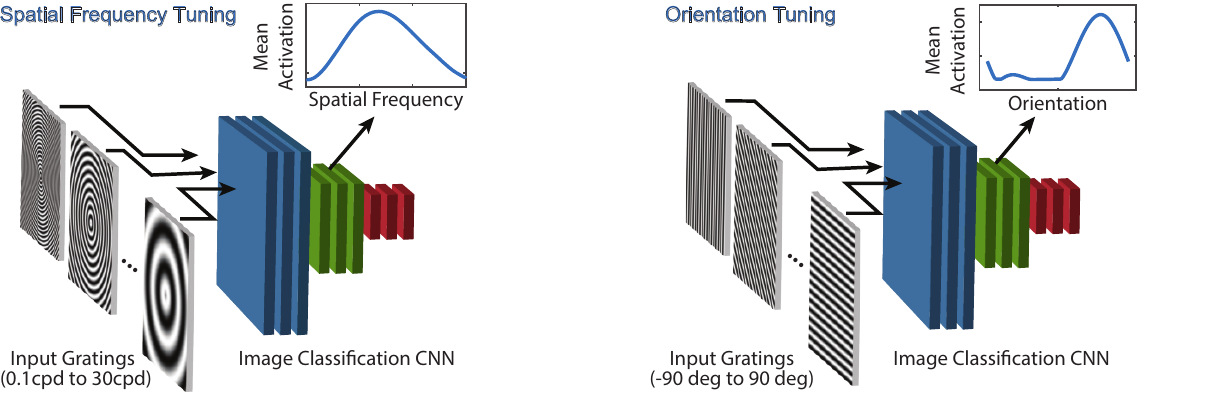}
\caption{Experimental setup: A pre-trained network is stimulated by gratings of varying spatial frequency (left) and gratings of fixed spatial frequency by varying orientation (right). We analyze the mean activations and derive a measure of spatial frequency and orientation tuning for each channel.}
\label{fig:block_dia}
\end{figure*}

\section{Perceptual Efficacy of Deep Features}
\label{sec:method}

Our method is inspired by the grating stimulus experiments traditionally used by neuroscientists to investigate the spatial frequency and orientation tuning dynamics in the HVS \cite{19}. Those experiments are based on the observation that a particular spatial frequency and orientation of a visual signal will elicit a spike in the neural activation of the visual cortex. We follow a similar approach and measure the spatial frequency selectivity and orientation tuning of different channels in activation maps of pre-trained deep CNNs.

\subsection{Inputs}
\label{fror_quant}

To quantify the spatial frequency tuning, we generate sinusoidal gratings of a fixed contrast and varying spatial frequencies. We denote these gratings by $I_f$ where $f$ is the grating frequency and use them to stimulate pre-trained CNNs. Then, we compute the spatial mean of the activation maps for each network channel as a function of the spatial frequency of the input signal. In this step, it is crucial to isolate the measurements from the effects of spatial orientation. To this end, we used radially symmetric grating patterns as our inputs. To quantify orientation selectivity, we repeat the process with a set of differently oriented sinusoidal gratings, denoted by $I_\theta$ where $\theta$ is the orientation. The gratings have a fixed spatial frequency that is selected as the peak of the CSF. Fig.~\ref{fig:block_dia} shows an overview of this analysis and sample input patterns. Analysing the response of CNN channels to gratings serves as a novel approach to visualize learned features, similar to \cite{Zeiler2013VisualizingAU,simonyan2014,erhan2009}.

\subsection{Measurement of the Spatial Frequency Sensitivity}
\label{vfss}
The previous investigations on the early stages of the human visual cortex show a behavior resembling a spatial frequency analyzer \cite{14}. Therefore, a significant portion of human visual perception is driven by the spatial frequency content of stimuli. The importance of understanding the effects of spatial contrast on low-level vision leads to one of the most widely studied models of the HVS, known as the contrast sensitivity function. The CSF depicts the HVS capability of perceiving contrast changes as a function of spatial frequency. Human observers have lower contrast discrimination thresholds at the spatial frequencies where the CSF reaches a high value (typically between $6-8$ cycles per degree). This corresponds to a higher probability of perceiving distortions, which contain spatial frequencies for which the contrast sensitivity function is high.


In our analysis, we assume that channels that exhibit higher sensitivity to changes in the spatial frequency are more likely to detect visual distortions since these usually change the spatial frequencies characteristic of an image. Additionally, the channels sensitive to perceptually-relevant distortions should follow the characteristic of the CSF. Consequently, we hypothesize that channels that are good for optimizing perceived quality are those which have high frequency sensitivity for the frequencies with a high CSF value.
We quantify this property of activation map channels by $\mu_1$ score:
\begin{equation}
\label{eq:2}
\mu_1(k,m) = \sum\limits_{f} CSF(f) \cdot \left| \frac{\partial }{\partial f} a_m^k (I_f) \right|,
\end{equation}
where $k$ is the index for the convolution layer, $m$ is the feature map index in each convolution layer, $CSF(\cdot)$ is the contrast sensitivity function, $a_m^k$ is the mean activation of the feature map and $f$ is the spatial frequency in cycles per degree.
$\mu_1$ quantifies the average frequency sensitivity of a CNN channel weighted by the CSF over different spatial frequencies. Channels with higher $\mu_1$ values should deliver better perceptual features according to our hypothesis. 
 
\begin{figure*}[t!]
\centering
\includegraphics[]{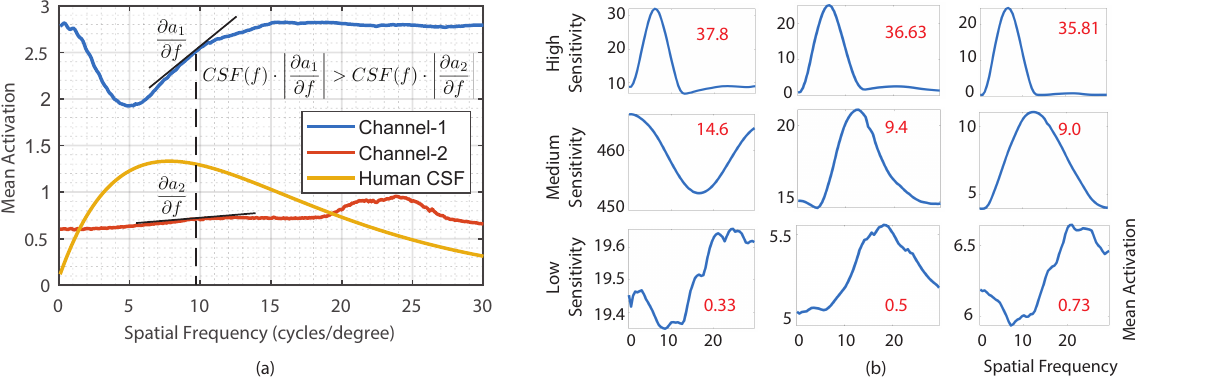}\label{fig:f7}
\caption{(a) Frequency sensitivity (b) Examples of the varying $\mu_1$ for channels in the \textit{ReLU2\_2} layer of the VGG-16 (denoted in red).}
\label{fig:freq_sen}
\end{figure*}

In Fig.~\ref{fig:freq_sen}(a), we provide mean activation plots of two channels from a deep CNN as a function of input spatial frequency. We observe that Channel-1 has a higher frequency sensitivity compared to Channel-2 at the frequencies where the CSF reaches its peak. From a perceptual perspective, Channel-1 has the desired behavior of responding to spatial frequency changes where the HVS has a higher sensitivity. Fig.~\ref{fig:freq_sen}(b) shows sample frequency tuning characteristics and the resulting values of $\mu_1$.


\subsection{Measurement of the Orientation Selectivity}
\label{orselec}

In addition to the underlying spatial frequency, orientation also plays an important role in the perception of visual stimuli. Previous studies indicate that the HVS presents orientation selectivity in the primary visual cortex for structure representation \cite{rani95,fester00}. Motivated by this fact, orientation selectivity is also a desired property for activation maps to detect artifacts in the form of structural deformations in visual stimuli.



We measure the orientation selectivity of activation maps by our quantitative score, $ \mu_2 $, which is based on the average squared difference between the activation and its peak value across all orientations of the input stimulus. For a channel \textit{m} in layer $\textit{k}$, the maximum activation with respect to input gratings of varying orientations $\theta$ can be calculated as:

%
\begin{equation}
    \hat{a}^k_m = \max_\theta {a^k_m ( I_\theta )}.
\end{equation}
Based on $\hat{a}^k_m$, we define our orientation selectivity score $ \mu_2 $ as:
%
\begin{equation}
\label{eq:3}
\mu_2(k,m) = \sum_\theta{ \left( a^k_m (I_\theta) - \hat{a}^k_m \right)^2}.
\end{equation}

\begin{figure*}[t!]
\centering
\includegraphics[]{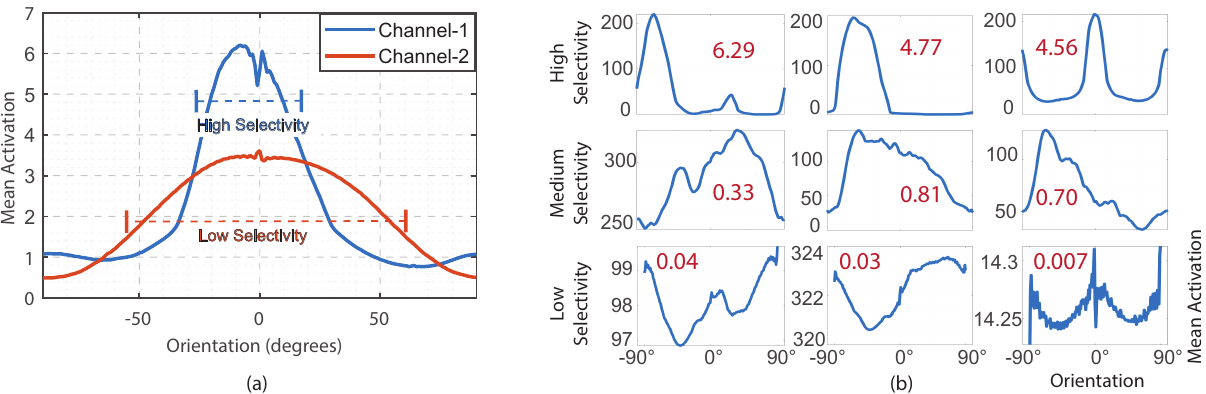}
\caption{(a) Orientation Selectivity (b) Examples of the varying $\mu_2$ for channels in the \textit{ReLU4\_2} layer of the VGG-16 (denoted in red).}
\label{fig:or_sel}
\end{figure*}

The above score resembles the statistical measure of kurtosis, but it is more efficient to compute for neural networks because it does not require the computation of standardized moments. Using Eq.~\ref{eq:3}, we compute the orientation selectivity of different channels in a pre-trained deep CNN from the inputs described in Sec.~\ref{fror_quant}. In Fig.~\ref{fig:or_sel}(a), the orientation selectivity characteristics of two selected sample channels from a network layer are shown. We observe that Channel-1 has a more significant orientation selectivity around $ 0^{\circ} $ compared to Channel-2. The higher selectivity of Channel-1 makes it a better candidate for detecting structural deformations which are visible to human observers. Fig.~\ref{fig:or_sel}(b) shows how mean activations change for a selected subset of channels from a deep CNN with respect to the stimulus orientation and the corresponding orientation selectivity scores computed using Eq.~\ref{eq:3}.


\subsection{Perceptual Efficacy (PE)}
\label{pe}

We combine our feature and orientation selectivity scores from Eq.~\ref{eq:2} and \ref{eq:3} into a single scalar representing the overall goodness of a feature channel for measuring and optimizing the perceptual quality. We call it the Perceptual Efficacy ($PE$). The perceptual efficacy of a channel with index \textit{m} in layer \textit{k} is the product of normalized $\mu_1$ and $\mu_2$:

%
\begin{equation}
\label{eq:4}
PE(\Phi^k_m) = \frac{\mu_1(k,m)\cdot\mu_2(k,m)}{\sum_m{\mu_1(k,m)}\cdot\sum_m{\mu_2(k,m)}}.
\end{equation}

\section{Experiments}
\label{setup}
We conduct a set of validation experiments and show that the feature channels with higher $PE$ scores have better overall correlation with subjective human quality judgments than the channels with lower $PE$ scores. In our analysis, the set of feature channels $F^k$, from layer $k$ of a pre-trained CNN is:
\begin{equation}
    F^k = \{\Phi^k_0, \Phi^k_1, \dots, \Phi^k_M \}.
\end{equation}

We split $F^k$ into subsets which consist of the channels with the highest and lowest $PE$ scores, denoted by $H^k \subseteq F^k$ and $L^k \subseteq F^k$, respectively. Furthermore, we control the cardinality of these subsets by changing the number of channels in each set based on percentile rank of channel $PE$ score. Using the ranking of channels according to their $PE$ scores, the set which consists of the channels with the top $x\%$ of $PE$ scores is:
\begin{equation}
    H^k\mbox{-}x = \{\Phi^k_i | PE(\Phi^k_i) \geq prc^k_{100-x} \}.
\end{equation}

Similarly, the set of channels with the lowest $PE$ scores is:
\begin{equation}
    L^k\mbox{-}x = \{\Phi^k_i | PE(\Phi^k_i) \leq prc^k_x \},
\end{equation}
where $prc^k_x$ is the $x^{th}$ percentile of $PE$ scores in layer $k$. For brevity, we omit the superscript $k$ when the subsets of channels in an experiment belong to the same layer of the network. To validate our hypotheses that lead to the formulation of $PE$, we compare the performance of channels in $H^k$ and $L^k$ in different tests, which are commonly used by the previous studies to measure the correlation with human perceptual quality assessment.





\subsection{Quality Assessment (QA) Tests}
\label{oqa}

\begin{figure*}[t]
\centering
\includegraphics[]{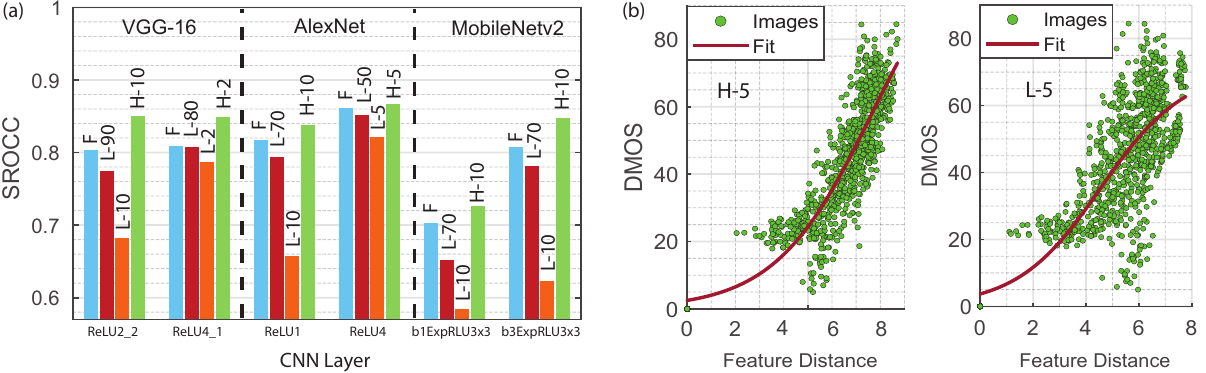}
\caption{Results of Quality Assessment (QA) test. (a) Spearman's correlation (SROCC) between human quality judgements (DMOS) and different subsets of feature channels ($F$, $H$ and $L$) from various deep CNNs. A higher SROCC value represents a performance closer to the human quality judgments. (b) Regression analysis between DMOS values and losses estimated by two subsets of feature channels, $H\mbox{-}5$ and $L\mbox{-}5$, from GoogleNet. We observe that smaller subsets of feature channels (denoted by $H$) selected according to our $PE$ score achieve a better SROCC and a better fit in DMOS regression.}
\label{fig:oqa}
\end{figure*}

QA tests are one of the most widely used techniques for benchmarking perceptual quality metrics. They aim to compute the correlation between the quality metrics and human subjective quality assessment scores called Differential Mean Opinion Scores (DMOS) \cite{24}. DMOS is a quantitative representation of how human observers perceive perceptual differences between natural and distorted images, and they are collected by conducting subjective experiments in a controlled environment where the observers evaluate varying levels of different distortion types. The performance of perceptual quality metrics is evaluated by computing statistics such as the RMSE (Root Mean Square Error), LCC (Linear Correlation Coefficient) and SROCC (Spearman Rank Order Correlation Coefficient) between the metrics and DMOS. Lower RMSE or higher LCC and SROCC indicate a better correlation with human perception of differences. Here, we report only the SROCC for simplicity. For a more detailed analysis and tests including all three evaluation metrics, please refer to the supplementary material.

For the purpose of this test, we compute the difference between images using the definition of perceptual loss in Eq.~\ref{eq:1} with different subsets of channels, as defined in Sec.~\ref{setup}. We use the images and DMOS scores from both the LIVE image quality dataset \cite{24} and multiple distortion dataset \cite{38} as our inputs. The two datasets include the distortions commonly observed in image processing applications such as Gaussian Blur, JPEG compression, JPEG2000, and White Noise, as well as combinations of these distortions. 

We hypothesize that subsets of channels in $H^k$ have a better correlation with DMOS compared to the channels in $L^k$. In order to show that this generalizes across different network architectures, we conduct the tests on different layers of several pre-trained image classification CNNs such as AlexNet, ShuffleNet, SqueezeNet, and VGG-16. The results (Fig.~\ref{fig:oqa}) demonstrate that indeed, higher correlations are achieved for the sets of channels with higher $PE$ scores.

\subsection{Just Noticeable Difference (JND) Test}
\label{jnd}

\begin{figure*}[t]
\centering
\includegraphics[]{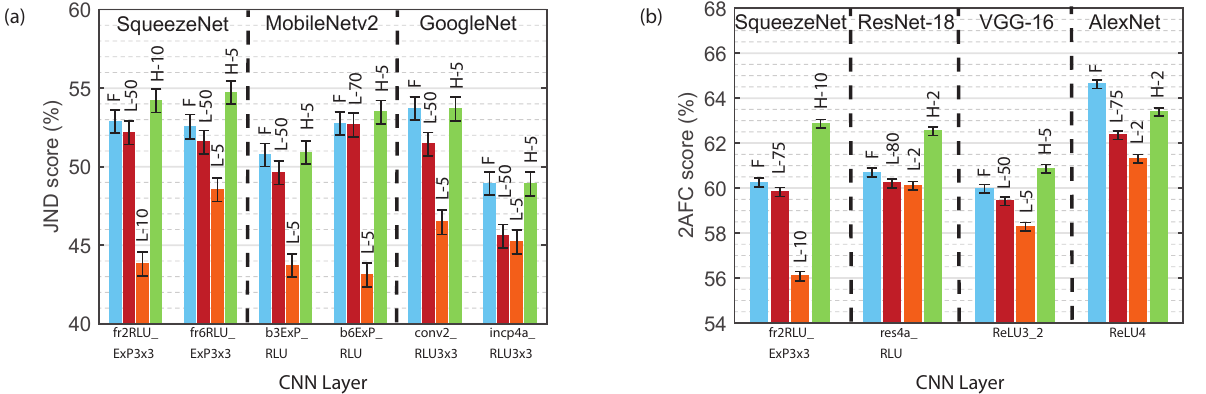}
\caption{JND (a) and 2AFC (b) test scores for different subsets of  feature channels selected according to their $PE$ scores. The sets of channels with a higher score are closer to human subjective quality judgments. Error bars represent Standard Deviation (SD). Please refer to Sec.~\ref{setup} for the definition of the channel subsets denoted by $F$, $L$ and $H$.}
\label{fig:bapps}
\end{figure*}

The Berkley-Adobe Perceptual Patch Similarity Dataset (BAPPS) is a perceptual similarity dataset which consists of image pairs and measures of Just Noticeable Differences between them \cite{12}. In the study conducted using this dataset, human observers were asked to determine whether a pair of images were perceptually the same or different. The dataset includes some distortions like spatial translation which are not represented in the other datasets that we used. For each pair of images, the study includes responses from three different observers who make a binary decision (same or different). If the difference between two images is below the detection threshold of observers, the responses tend to be in agreement as ``same''. For each observer, the net score for each pair of images is represented by a rational number that can be either 0 (consensus on ``same''), 1 (consensus on ``different''), 1/3 (one out of the three reports ``same'') or 2/3 (one out of the three reports ``different''). In this test, perceptual metrics should be able to detect the distortions visible to humans and successfully assign the images into the four classes represented by the rational scores. Similarly to the QA test, we compute the difference between image pairs using different subsets of feature channels selected according to our $PE$ scores. We then compute the JND score as a percentage of images correctly classified (Fig.~\ref{fig:bapps}(a)). We conduct this test using the CNN-based distortion set, which is highly relevant to the deep learning solutions but not represented in QA tests. The results show that, also, for this test, our $PE$ measure is beneficial.


\subsection{2AFC Similarity Tests}
\label{2afc}
In the 2AFC test, two distorted images ($I_1$, $I_2$) and a reference image $I_{\text{GT}}$ are shown to observers who are asked to choose the distorted image that is closer to the reference \cite{12}. Perceptual metrics are evaluated by measuring their agreement with the pair-wise human judgments as follows:
Scores from a perceptual metric are converted to binary responses ($I_1$ or $I_2$) depending on the distance of those images to $I_{\text{GT}}$, resembling the binary decisions of human observers. Assuming that the fraction $p$ of the observers choose $I_1$ and ($1-p$) choose $I_2$, the perceptual metric is assigned a ``credit'' of $p$ if the metric indicates $I_1$ and $(1-p)$ otherwise. As a result of this process, the perceptual metric that successfully chooses more popular images among human observers accumulates a higher amount of credit indicating their performance. In these tests we again use the BAPPS dataset and the computed scores are shown in Fig.~\ref{fig:bapps}(b).

\subsection{Visual Evaluation of the Features}
\label{vis_invert}

\begin{figure*}[t]
\centering
\includegraphics[]{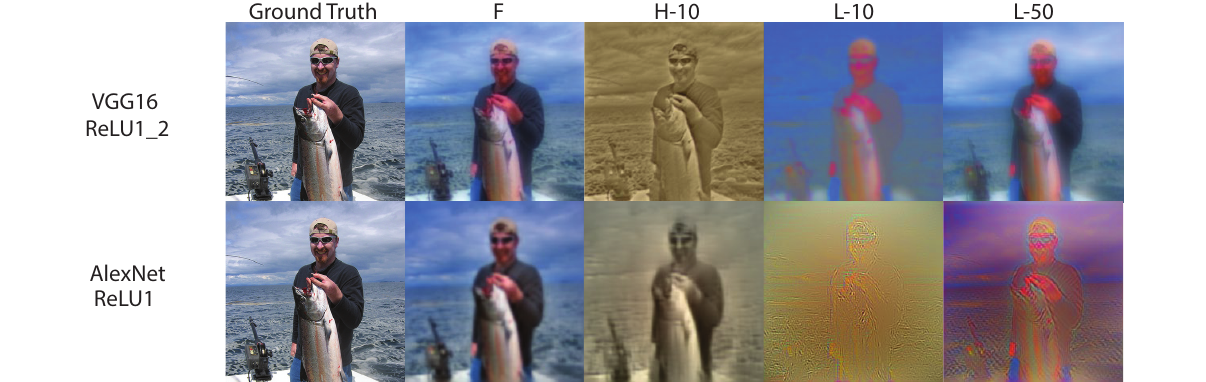}
\caption{Visualization of the information encoded in different sets of feature channels with high ($H\mbox{-}10$) and low ($L\mbox{-}10$ and $L\mbox{-}50$) $PE$ scores.}\label{fig:invert}
\end{figure*}

To visualize the image information encoded at different layers of a CNN, it is possible to reconstruct an image from a set of CNN features \cite{48}. We apply this method to investigate the information encoded by different sets of feature channels with high and low $PE$ scores (Fig.~\ref{fig:invert}). We observe that the channels with high $PE$ scores encode information that can be considered more visually relevant, e.g., edges, while color is encoded in the channels with lower $PE$ scores.

\subsection{Super-Resolution}
\label{sr}

\begin{figure*}[t]
\centering
\includegraphics[]{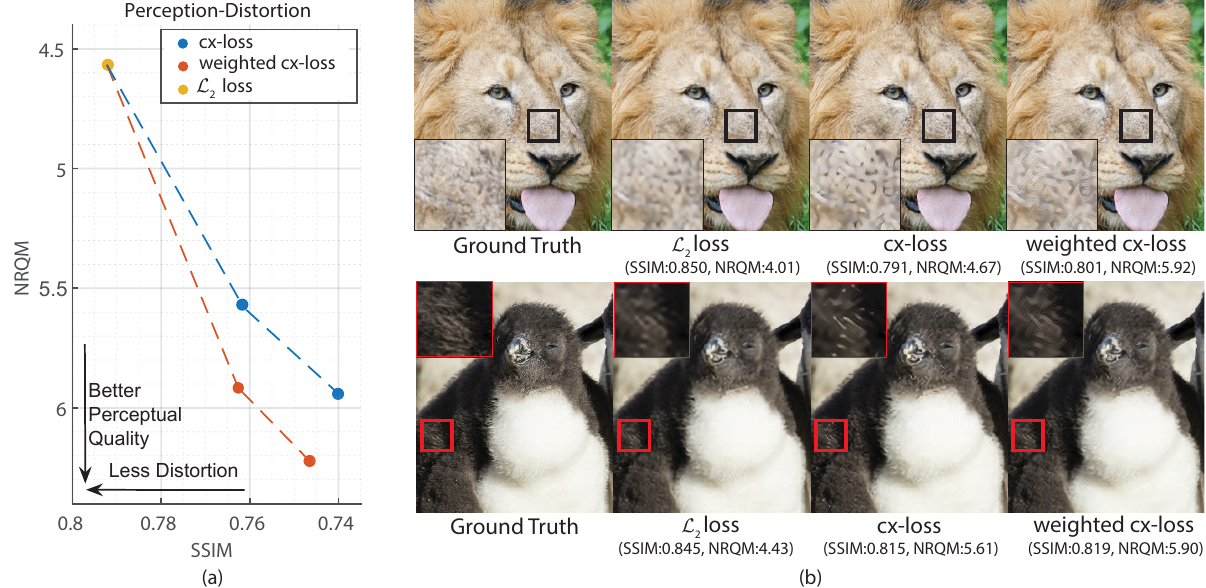}
\caption{Results from the Super-Resolution (SR) experiment. (a) shows the change in NRQM \cite{31} and SSIM \cite{6} for $\mathcal{L}_2$ and contextual (cx) loss. Weighting the channels used in contextual loss proportionally to their $PE$ scores moves the network towards a more optimal point in the perception-distortion plane represented by NRQM and SSIM metrics. In (b) we provide sample outputs from the networks trained with different losses (SR scale factor: $\times4$).}
\label{fig:sr_results}
\end{figure*}

We apply the results of our analysis to the widely used contextual loss for super-resolution (SR) \cite{32}. The contextual loss (cx-loss) is known for its better perceptual quality in SR applications compared to the perceptual loss of Johnson~et~al.~\cite{7}. We observe that weighting feature channels according to $PE$ is a promising approach for improving contextual loss performance. We perform an $\times4$ SR experiment using the VDSR~\cite{26} network, trained on the images of the DIV2K~\cite{27} dataset. The loss function used was a combination $\mathcal{L}_2$ loss and the contextual loss (using \textit{ReLU2\_2} of the VGG-16). 
\begin{equation}
\label{eq:sr_loss}
    \mathcal{L}_{\text{total}} = \alpha\cdot \mathcal{L}_2 + (1-\alpha)\cdot \mathcal{L}_{\text{contextual}}
\end{equation}
The recent work of Blau~and~Michaeli~\cite{5} has drawn attention to the presence of a tradeoff between the improvements in perceptual quality and pixel-wise error measures in image reconstruction tasks. Recently some approaches such as the adversarial training started providing the flexibility of controlling this tradeoff during training by carefully tuning the weights used in their loss functions. In Eq.~\ref{eq:sr_loss}, it is possible to move along this tradeoff boundary by changing $\alpha$. Since it is trivial to move along this line in the perception-distortion plane without changing the underlying reconstruction method, it has become necessary to use multiple metrics that evaluate both perceptual and low-level visual quality aspects to accurately evaluate the image quality. Motivated by this observation, we evaluated the reconstruction quality of several networks with different perception-distortion tradeoff. We additionally considered weighting the features used in the loss function according to our $PE$ scores. We observe that the weighting improves the results according to both perceptual quality and pixel-wise distortion metrics (Fig.~\ref{fig:sr_results}).

\subsection{Discussion}
\label{sec:discussion}

The Quality Assessment (QA), Just Noticeable Difference (JND) and 2AFC validation experiments that we conducted show that the loss functions which use the feature channels with the highest $PE$ scores have better correlation with human quality judgments. This proves our hypothesis on the deep feature representations and it also shows that even small subsets of channels carefully selected according to higher $PE$ scores are successful at picking up just noticeable distortions and they may easily outperform the complete set of channels in a layer. We believe that leaving out the feature channels with lower $PE$ scores reduces the redundancy in the optimization by focusing on the visual properties which are more dominant in driving human quality perception.

Visual evaluation of the information encoded in different subsets of feature channels also supports this observation from QA, JND and 2AFC tests. The result of this analysis on two different layers of VGG16 and AlexNet in Fig.~\ref{fig:invert}, leads us to two interesting observations. First, we see that the channels with high $PE$ scores ($H\mbox{-}10$) encode the medium spatial frequencies which are closer to the peak of the CSF. Those frequencies are essential to recover important details of natural textures and image statistics in image reconstruction tasks. Second, the visual information encoded in the channels with high $PE$ scores represents achromatic contrast whereas the channels with the lowest $PE$ scores ($L\mbox{-}10$) focus more on color. This observation is also in agreement with previous psychophysical experiments which show that HVS is better at discriminating luminance details compared to color \cite{mantiuk2013}. We believe that these two properties of feature channels have important implications on the sensitivity of these features for detecting visible distortions.

Fig.~\ref{fig:sr_results} shows that weighting the channels of the contextual loss according to $PE$, is a promising approach for improving perception-distortion image characteristics. We observe the tradeoff between perceptual quality and distortion by changing the $\alpha$ parameter in Eq.~\ref{eq:sr_loss}. Weighting the feature channels according to their $PE$ scores improves overall quality by moving the tradeoff line towards the origin in the perception-distortion plane. Using only $\mathcal{L}_2$ loss by setting $\alpha = 1$ results in the poorest NRQM \cite{31} score as expected because $\mathcal{L}_2$ does not measure the perceptual quality. In Fig.~\ref{fig:sr_results}(b), we provide some sample images from this experiment where the quality improvement obtained by weighting the cx-loss is visible in the sample images. We believe that our analysis and showing potential improvement in well-known perception-distortion tradeoff will inspire further investigations in this promising research direction.

\section{Comparison with LPIPS and Other Metrics}
\label{sec:comparison}
\begin{figure*}[h]
\centering
\includegraphics[]{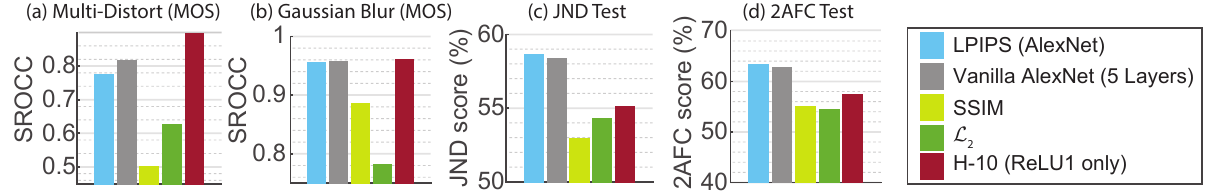}
\caption{The correlation of metric scores with MOS for images corrupted with a combination of multiple (a) and single (b) distortions from the LIVE dataset. (c) JND test scores for CNN distortions. (d) 2AFC scores for frame-interpolation.}
\label{fig:lpips_res}
\end{figure*}

We also evaluate how our strategy for selecting features improves the prediction of perceptual data when compared to other methods. In Fig.~\ref{fig:lpips_res}, we compare the performance of our $H\mbox{-}10$ to LPIPS, original AlexNet, SSIM, and $\mathcal{L}_2$ using different tests. Our method outperforms the last two methods in all tests. Below, we focus our discussion on the comparison to LPIPS, which is a full-reference perceptual quality measure based on pre-trained CNN representations with many successful applications \cite{43,44,45,46,47}. 

First, we perform two QA tests, using images and DMOS scores of the LIVE multi-distortion dataset with multiple types of distortions which are sometimes combined, making the distortions difficult to model. Combining distortions increase the sample diversity that are unseen by LPIPS during training. We argue that this is a fair comparison since our technique does not require training. Second, we perform a QA test with images with a much simpler distortion (Gaussian blur). We employ AlexNet LPIPS framework as it has the best performance. Additionally, we also include JND and 2AFC tests, for which LPIPS was trained.

Fig.~\ref{fig:lpips_res}(a) shows that LPIPS might not always have the best correlation with MOS, especially for the distortions that are not represented in the training set. The vanilla five-layer AlexNet framework performs better in that setting. Most interestingly, using only 10\% of good channels ($H\mbox{-}10$) from a single layer of a pre-trained AlexNet demonstrates a higher correlation with human opinions. Fig.~\ref{fig:lpips_res}(b) shows that for images with Gaussian blur, the vanilla AlexNet still has slightly better correlation with human MOS compared to LPIPS, and even 10\% of good channels from the \textit{ReLU1} layer can perform better. For JND and 2AFC tests LPIPS outperforms other uncalibrated metrics (Fig.~\ref{fig:lpips_res}(c) and (d)).

The better performance of LPIPS on JND and 2AFC tests is expected since the metric was trained for these tasks. On the other hand, we believe that its lower performance on MOS dataset is because it relies on pair-wise judgments of similarity with the reference images, which may not capture how differently humans perceive the transitions in distortion levels. This is important because, in some cases, MOS-based experiments may be preferred to analyze image quality \cite{24}. The experiment also shows that feature selection is critical for the design of an effective perceptual loss. In some cases, features trained specifically on perceptual data can be outperformed by carefully selected features from a network trained for a different task, e.g., classification. To summarize, the conclusions to be drawn from this analysis are:
\begin{enumerate}
\item LPIPS is very good for patch-based similarity but has room for improvement in terms of overall correlation with MOS, especially on novel distortions.
\item Selection of the feature space plays a very important role in how well feature distances correlate to human MOS, using as many features as possible does not necessarily correlate with better performance. 
\item Is it beneficial to integrate good feature selection with a modified training methodology to create metrics that correlate well with both human MOS and patch-wise judgments. Analysis of the implications of MOS-based and patch-based tests is left as an open problem for future work.
\end{enumerate}


\section{Conclusion}
In this work, we explored feature characteristics of deep representations and compared them with fundamental aspects of HVS using our frequency sensitivity and orientation selectivity measures. We showed that the features selected according to higher CSF-weighted frequency sensitivity and higher orientation selectivity exhibit a higher correlation with human quality judgments. 

The general practice when training neural networks is using the full set of available feature channels from a pre-trained deep CNN. Our findings suggest that a more optimal solution is to focus the loss function on a subset of features which are ranked high using our measures. We validated this hypothesis on an example of super-resolution by re-weighting the features according to their frequency sensitivity and orientation selectivity. We demonstrated that such a strategy achieves better perception-distortion trade-off.

We believe that our analysis is an essential step towards better understanding of the connection between human visual perception and learned features. Furthermore, our work opens new research venues by laying a foundation for analyzing the efficiency of deep feature representations. Extending our analysis by more advanced aspects of perception, such as visual masking or temporal sensitivity, is an exciting direction for future work which may lead to better understanding of deep representations and how they relate to human perception. It is also possible to apply similar feature analysis and optimizations to other domains with established models of human perception, e.g., audio processing. 

Due to the growing popularity of deep learning in different domains, we anticipate seeing more improvements in perceptual quality optimization as the connection between HVS and learned features is established.

\subsubsection*{Acknowledgements:}
This project has received funding from the European Research Council (ERC) under the European Union’s Horizon 2020 research and innovation program (grant agreement N$^{\circ}$ 804226 – PERDY).


\clearpage
%
%
\bibliographystyle{splncs04}

\end{document}